\pdfoutput=1

\documentclass[11pt]{article}

\usepackage{EMNLP2022}

\usepackage{times}
\usepackage{latexsym}

\usepackage[T1]{fontenc}

\usepackage[utf8]{inputenc}

\usepackage{microtype}

\usepackage{inconsolata}

\usepackage{color}
\usepackage{booktabs}
\usepackage{graphicx}
\usepackage{xcolor, colortbl}
\usepackage{subfigure}
\usepackage{array}
\usepackage{multirow}
\usepackage{multicol}
\usepackage{adjustbox}
\usepackage[english]{babel}

\title{How Hate Speech Varies by Target Identity: A Computational Analysis}

\author{Michael Miller Yoder\textmd{\textsuperscript{1}} \hspace*{.4ex} Lynnette Hui Xian Ng\textmd{\textsuperscript{1}} \hspace*{.4ex} David West Brown\textmd{\textsuperscript{2}} \hspace*{.4ex} Kathleen M. Carley\textmd{\textsuperscript{1}} \\
    \textsuperscript{1}School of Computer Science \hspace*{0.5ex} \textsuperscript{2}Department of English \\
    Carnegie Mellon University \\
    Pittsburgh, Pennsylvania, USA \\
  \texttt{\{mamille3,huixiann,dwb2,carley\}@andrew.cmu.edu} \\
  }

\begin{document}
\maketitle
\begin{abstract}
This paper investigates how hate speech varies in systematic ways according to the identities it targets.
Across multiple hate speech datasets annotated for targeted identities, we find that classifiers trained on hate speech targeting specific identity groups struggle to generalize to other targeted identities.
This provides empirical evidence for differences in hate speech by target identity; we then investigate which patterns structure this variation.
We find that the targeted demographic category (e.g. gender/sexuality or race/ethnicity) appears to have a greater effect on the language of hate speech than does the relative social power of the targeted identity group.
We also find that words associated with hate speech targeting specific identities often relate to stereotypes, histories of oppression, current social movements, and other social contexts specific to identities.
These experiments suggest the importance of considering targeted identity, as well as the social contexts associated with these identities, in automated hate speech classification. 

\textit{\textbf{Warning:} This paper contains offensive and hateful terms and concepts.
We have chosen to reproduce these terms for clarity in aiding efforts against hate speech.}
\end{abstract}

\section{Introduction}

Researchers working in natural language processing (NLP) often treat hate speech as a binary, unified, concept that can be detected from language alone.
However, as a linguistic concept that relies heavily on social context, hate speech contains a variety of related phenomena~\citep{Brown2017part2}. 
Hate speech is characterized by variation in linguistic features (e.g. implicit vs. explicit), context (e.g. platforms, prior conversations), and communities (social histories and hierarchies). 
This paper focuses on a crucial aspect of this variation: how hate speech varies by the identity groups it targets. 

To study this variation, we analyze hate speech datasets that include annotations for which identity group is targeted.
Drawing from multiple of these datasets, we sample new corpora that target the same identity group.
These identity groups vary according to several dimensions, including relevant demographic category (e.g. gender, religion) and relative social power (e.g. socially marginalized or dominant).
We empirically test which dimensions most clearly separate different forms of hate speech by
evaluating how well classifiers trained on one set of identities generalize to hate speech directed at different sets of identities. 

We find that hate speech varies most prominently by the targeted demographic category and less so by the social power of the targeted identity group. 
Theorists working in philosophy and sociolinguistics have drawn attention to how hate speech directed at marginalized groups differs from hate directed toward socially dominant groups \citep{Butler1997, Lakoff2000}.
However, we do not find that hate speech toward dominant groups is sufficiently different to consistently increase classification performance when removed from existing datasets.

Analyzing the most representative terms in hate speech directed toward different identities, we find that many words reflect identity-specific context such as histories of oppression or stereotypes.
These results have implications for NLP researchers building generalizable hate speech classifiers, as well as for a more general understanding of variation in hate speech.

\paragraph{Contributions}
\begin{enumerate}
    \item An empirical analysis of variation in hate speech by target identity. Specifically, how well classifiers trained on hate speech directed toward specific identities generalize to hate speech directed at other identities.
    \item An analysis of which dimensions of social difference (demographic category, power) among targeted identities  reflect the most variation in hate speech.
    \item A qualitative analysis of the hate speech terms most strongly associated with specific target identities.
\end{enumerate}

\section{Hate Speech}
Hate speech is an example of a ``thick concept'' with a set of related, but difficult to define meanings and understandings \citep{Pohjonen2017}.
Legal theorist Alexander Brown \citeyearpar{Brown2017part2} argues for a set of attributes that make an expression more or less likely to be considered hate speech, similar to Wittgenstein's ``family resemblances'' concept.
Key attributes include an incitement of emotion and violence, and a direction of that incitement toward a targeted identity group \citep{Sanguinetti2018, Poletto2021}.
Though others have studied the linguistic properties of this incitement~\citep{Marsters2019,wiegand-etal-2021-implicitly-abusive}, we focus on how variation in the identity group targeted by hate speech affects the linguistic characteristics of hate speech.


\subsection{Variation by identity}
Identities are central to hate speech.
Classifiers often learn to associate the presence of identity terms, especially derogatory ones, with hate speech and abusive language~\citep{Dixon2017,Uyheng2021}.
Computational studies of the targets of online hate speech have included measurement studies of its prevalence toward different targets.
\citet{Silva2016} and \citet{Mondal2017} searched for templates such as ``I hate \_\_\_'' to measure hate toward different identity groups.
We analyze datasets manually annotated with the targets of hate speech.
This captures a broader range of hate speech, including indirect hate speech and stereotypes.
\citet{elsherief2018hate,ElSherief2018} investigated differences between hate toward groups versus individual targets.
In contrast, we compare differences among identity targets.
\citet{rieger_assessing_2021} measured multiple types of variation, including by identity target, in hate speech from fringe platforms such as 4chan and 8chan.
We test if such differences affect the generalization of hate speech classifiers.

Many identities are involved in the production and recognition of hate speech, including the identities of those who produce hate speech and those who annotate hate speech datasets.
The post history and inferred gender of social media users have been found to be useful in predicting hate speech~\citep{Waseem_hovy2016,unsvag_effects_2018,qian_leveraging_2018}.
\citet{Waseem2016} find differences in hate speech annotations between crowdworkers and experts, while \citet{sap2022annotators} find differences by the political ideology of annotators.
We focus on identities presented in the hate speech itself.

\subsection{Generalizability}
In this paper, we evaluate the ability of hate speech classifiers to generalize across targeted identities.
\citet{Groendahl2018} find that hate speech models generally perform poorly on data that differs from their training data; we look at how shifts in the distribution of identity targets affects generalization.
\citet{Swamy2019} look at generalizability across subtasks of abusive language detection and find that a larger proportion of hateful instances aids generalization.
\citet{Pamungkas2020} and \citet{fortuna-etal-2020-toxic} find that hate speech models using variants of BERT \citep{devlin-etal-2019-bert} generalize better than other models.
We thus use a variant of BERT in our generalization experiments.
See \citet{Yin2021} for a more thorough survey on generalizability in hate speech detection.

\section{Data}
From surveys of hate speech datasets \citep{Vidgen2020,Poletto2021} and the Hate Speech Dataset Catalogue\footnote{\url{https://hatespeechdata.com/}}, we selected datasets with annotations for targeted identities.
We only selected datasets that do not restrict target identities in order to minimize differences in other properties (e.g, domain, year) when comparing across targeted identities.
This excludes hate speech datasets and shared tasks that focus on particular targeted identity groups, such as women or immigrants \citep{Kwok2013,Basile2019}.

We also did not consider hate speech datasets that label targeted demographic category, such as race or gender \citep{Waseem2016}, but do not specify the identity group targeted.
Demographic category is just one of the dimensions of similarities and differences among identity groups that we wish to compare for their affect on hate speech.
We included datasets from all domains, except those with synthetic data.

Since we only found one non-English dataset that contained unrestricted annotations for targeted identities \citep{ousidhoum_multilingual_2019}, we focus on hate speech in English in this work.

For generalization analyses, we sampled corpora specific to identity groups across datasets large enough to contain a minimum number of instances of hate speech against enough groups (described in Section \ref{sec:sampling}).
These are the first 4 datasets noted in \autoref{tab:dataset}.
All datasets are used in the analysis of removing dominant groups (Section \ref{sec:power_removal_comparison}).
\begin{table*}[tb]
\centering
\begin{tabular}{llr}
\textbf{Dataset} & \textbf{Domain} & \textbf{Original size} \\  
\hline
Civil Comments~\citep{Borkan2019} & News comments & 1999516 \\  
Social Bias Inference Corpus~\citep{Sap2020} & Reddit, Twitter, Gab, Stormfront & 44781\\ 
\citet{Kennedy2020} & YouTube, Twitter, Reddit & 39565 \\  
HateXplain ~\citep{mathew2021hatexplain} & Twitter, Gab & 20148\\
\hline
Contextual Abuse Dataset~\citep{vidgen-etal-2021-introducing} & Reddit & 27494 \\  
\citet{elsherief2021latent} & Twitter & 19650 \\  
\citet{salminen2018anatomy} & YouTube, Facebook & 3222 \\  
\end{tabular}
\caption{Overview of datasets used in this study. Original size is the number of instances before resampling for experiments. The last 3 datasets are only used in the experiment removing hate toward dominant social groups (section \ref{sec:power_removal_comparison}).}
\label{tab:dataset}
\end{table*}

Datasets are resampled to a 30/70 ratio of hate to non-hate to eliminate a source of variance among hate speech datasets known to affect generalization~\citep{Swamy2019}.
Non-hate instances are upsampled or downsampled to meet this ratio, which was chosen as typical of hate speech datasets~\citep{Vidgen2020}.
If they do not already contain a binary hate speech label, dataset labels are binarized as described in Appendix \ref{sec:appendix}.

\subsection{Target identity label normalization}
\label{sec:norm}
Annotations for targeted identities vary considerably across datasets.
Some of these differences are variations in naming conventions for identity groups with significant similarity (`Caucasian' and `white people', for example).
Other identities are subsets of broader identities, such as `trans men' as a specific group within `LGBTQ+ people'.

To construct identity-based corpora across datasets, we normalized and grouped identities annotated in each dataset.
One of the authors, who has taken graduate-level courses on language and identity, manually normalized the most common identity labels in each dataset and assigned these normalized identity labels into broader identity groups (such as `LGBTQ+ people').
Intersectional identities, such as `Chinese women', were assigned to multiple groups (in this case `Asian people' and `women').
Hate speech was often directed at conflated, problematic groupings such as `Muslims and Arabs'.
Though we do not condone these groupings, we use them as the most accurate descriptors of identities targeted.

\section{Cross-Identity Generalization}
\label{sec:cross-identity}
We examine variation among hate speech targeting different identities in a bottom-up, empirical fashion.
In order to do this, we construct corpora of hate speech directed at the most commonly annotated target identities, grouped and normalized as described in Section \ref{sec:norm}.
We then trained hate speech classifiers on each target identity corpus and evaluated on corpora targeting other identities.

Along with practical implications for hate speech classification generalization, this analysis suggests which similarities and differences among identities are most relevant for differentiating hate speech.

\subsection{Data sampling}
\label{sec:sampling}
In order to have enough data targeting many identities and to generalize beyond the particularities of specific datasets, we assembled identity-specific corpora from multiple source datasets.
To mitigate dataset-specific effects, we uniformly sampled hate speech instances directed toward target identities from the first 4 datasets listed in \autoref{tab:dataset}.
We select these datasets since they contain enough data to train classifiers targeting a sufficient variety of identities.
The corpus for each target identity contains an equal amount of hate speech drawn from each of these datasets, though the total number of instances may differ among corpora.
Negative instances were also uniformly sampled across datasets, and were restricted to those which had no target identity annotation or an annotation that matched the target identity of the hate speech.

We selected target identities that contained a minimum of 900 instances labeled as hate across these four datasets after grouping and normalization.
We selected this threshold as a balance between including a sufficient number of identities and having enough examples of hate speech toward each identity to train classifiers.
In order to include a variety of identities in the analysis while maintaining uniform samples for each dataset, we upsample identity-specific hate speech from individual datasets up to 2 times if needed.
Corpora are split into a 60/40 train/test split.
Selected target identities and the size of each corpus can be found in \autoref{tab:corpora_stats}. These identity-specific corpora, which are samples of existing publicly available datasets, are available at \url{https://osf.io/53tfs/}.

\begin{table}[tb]
\centering
\begin{tabular}{lrr}
\bf Corpus & \bf Train size & \bf Test size \\
 \toprule
Women & 27960 & 18624 \\
Black people & 17664 & 11776 \\
Muslims, Arabs & 13712 & 9136 \\
LGBTQ+ people & 10544 & 7000 \\
Asian people & 7968 & 5312 \\
Latinx people & 7016 & 4688 \\
Jews & 5080 & 3400 \\
White people & 2328 & 1560 \\
Men & 1832 & 1232 \\
Christians & 1816 & 1224 \\
\midrule
Race/ethnicity & 71024 & 47240 \\
Gender/sexuality & 63032 & 42056 \\
Religion & 32144 & 21376 \\
\midrule
Marginalized & 168904 & 112792 \\
Dominant & 7952 & 5368 \\
\end{tabular}
\caption{Number of instances in corpora used in generalization experiments. 
These corpora are sampled by target identity uniformly from the first 4 datasets listed in \autoref{tab:dataset}.}
\label{tab:corpora_stats}
\end{table}

\subsection{Cross-identity hate speech classification}
Due to the high performance of BERT-based models on hate speech classification~\citep{Mozafari2019,Samghabadi2020}, we trained and evaluated a DistilBERT model~\citep{Sanh2019}, which has been shown to perform very similarly to BERT on hate speech detection with fewer parameters~\citep{vidgen-etal-2021-introducing}.
Models were trained with early stopping after no improvement for 5 epochs on a development set of 10\% of the training set.
An Adam optimizer was used with an initial learning rate of $10^{-6}$.
Input data was lowercased and an uncased base DistilBERT model was fine-tuned using the Hugging Face Transformers package, Keras, and Tensorflow.
We removed URLs, hashtags and @mentions of users, but kept emoji in preprocessing.
To mitigate random variation, we trained separate DistilBERT models 5 times and report the average performances.

As a baseline, we also evaluated a logistic regression classifier with TF-IDF unigram features over the entire vocabulary.
This classifier used L2 regularization with a constant $C=1$.

Results from only the DistilBERT models are reported as they consistently outperformed the logistic regression model by 0.1 F1 or more.
Generalization performance trends across identities were similar for DistilBERT and logistic regression.
Code for these analyses are available at \url{https://github.com/michaelmilleryoder/hate_speech_identities}.

\subsection{Results}
\autoref{fig:identity_generalization} shows generalization performance, measured by F1-score on the positive class of hate speech, across identity splits.
We choose F1 on the `hate' class since that focuses on performance in detecting hate speech across different target identities, rather than the non-hate instances which may or may not target identities.
Generalization across target identities is poor, often dropping from over 70 F1-score when training and test sets match by targeted identity to less than 40 when they do not.

\newcolumntype{M}[1]{>{\centering\arraybackslash}m{#1}}

\newcolumntype{R}[2]{%
    >{\adjustbox{angle=#1,lap=\width-(#2)}\bgroup}%
    l%
    <{\egroup}%
}
\newcommand*\rot{\multicolumn{1}{R{-40}{-1em}}}

\begin{table}[tb]
    \centering
\setlength\tabcolsep{0pt}
\tiny
\begin{tabular}{c@{\hspace{-1.2em}}r@{\hspace{0.7em}}@{\rule[-0.2cm]{0pt}{.5cm}}*{10}{M{.5cm}}}
\parbox[t]{5mm}{\multirow{19}{*}{\rotatebox[origin=c]{90}{\textbf{Train}}}} & Asian & {\cellcolor[HTML]{09529D}} \color[HTML]{F1F1F1} 71.5 & {\cellcolor[HTML]{97C6DF}} \color[HTML]{000000} 40.2 & {\cellcolor[HTML]{C7DBEF}} \color[HTML]{000000} 30.6 & {\cellcolor[HTML]{9CC9E1}} \color[HTML]{000000} 39.4 & {\cellcolor[HTML]{5DA5D1}} \color[HTML]{F1F1F1} 49.9 & {\cellcolor[HTML]{D9E7F5}} \color[HTML]{000000} 24.4 & {\cellcolor[HTML]{D3E3F3}} \color[HTML]{000000} 26.6 & {\cellcolor[HTML]{ADD0E6}} \color[HTML]{000000} 35.9 & {\cellcolor[HTML]{8ABFDD}} \color[HTML]{000000} 42.2 & {\cellcolor[HTML]{D7E6F5}} \color[HTML]{000000} 24.9 \\
& Black & {\cellcolor[HTML]{9CC9E1}} \color[HTML]{000000} 39.5 & {\cellcolor[HTML]{083776}} \color[HTML]{F1F1F1} 78.2 & {\cellcolor[HTML]{CADDF0}} \color[HTML]{000000} 29.7 & {\cellcolor[HTML]{BDD7EC}} \color[HTML]{000000} 32.7 & {\cellcolor[HTML]{65AAD4}} \color[HTML]{F1F1F1} 48.4 & {\cellcolor[HTML]{DAE8F6}} \color[HTML]{000000} 23.9 & {\cellcolor[HTML]{C7DCEF}} \color[HTML]{000000} 30.3 & {\cellcolor[HTML]{CDE0F1}} \color[HTML]{000000} 28.4 & {\cellcolor[HTML]{60A7D2}} \color[HTML]{F1F1F1} 49.3 & {\cellcolor[HTML]{CDDFF1}} \color[HTML]{000000} 28.6 \\
& Christians & {\cellcolor[HTML]{DBE9F6}} \color[HTML]{000000} 23.7 & {\cellcolor[HTML]{D1E2F3}} \color[HTML]{000000} 27.1 & {\cellcolor[HTML]{529DCC}} \color[HTML]{F1F1F1} 52.1 & {\cellcolor[HTML]{94C4DF}} \color[HTML]{000000} 40.5 & {\cellcolor[HTML]{D1E2F3}} \color[HTML]{000000} 27.1 & {\cellcolor[HTML]{D6E5F4}} \color[HTML]{000000} 25.4 & {\cellcolor[HTML]{DFECF7}} \color[HTML]{000000} 22.2 & {\cellcolor[HTML]{B9D6EA}} \color[HTML]{000000} 33.5 & {\cellcolor[HTML]{D6E5F4}} \color[HTML]{000000} 25.6 & {\cellcolor[HTML]{E2EDF8}} \color[HTML]{000000} 21.5 \\
& Jews & {\cellcolor[HTML]{E4EFF9}} \color[HTML]{000000} 20.6 & {\cellcolor[HTML]{E3EEF8}} \color[HTML]{000000} 21.2 & {\cellcolor[HTML]{B2D2E8}} \color[HTML]{000000} 35.0 & {\cellcolor[HTML]{08306B}} \color[HTML]{F1F1F1} 79.9 & {\cellcolor[HTML]{EBF3FB}} \color[HTML]{000000} 18.3 & {\cellcolor[HTML]{EDF4FC}} \color[HTML]{000000} 17.7 & {\cellcolor[HTML]{F6FAFF}} \color[HTML]{000000} 14.8 & {\cellcolor[HTML]{D6E5F4}} \color[HTML]{000000} 25.5 & {\cellcolor[HTML]{E1EDF8}} \color[HTML]{000000} 21.7 & {\cellcolor[HTML]{F7FBFF}} \color[HTML]{000000} 14.3 \\
& Latinx & {\cellcolor[HTML]{7CB7DA}} \color[HTML]{000000} 44.5 & {\cellcolor[HTML]{9CC9E1}} \color[HTML]{000000} 39.4 & {\cellcolor[HTML]{B9D6EA}} \color[HTML]{000000} 33.4 & {\cellcolor[HTML]{AFD1E7}} \color[HTML]{000000} 35.5 & {\cellcolor[HTML]{125EA6}} \color[HTML]{F1F1F1} 68.2 & {\cellcolor[HTML]{DAE8F6}} \color[HTML]{000000} 24.1 & {\cellcolor[HTML]{DCEAF6}} \color[HTML]{000000} 23.2 & {\cellcolor[HTML]{C8DCF0}} \color[HTML]{000000} 30.1 & {\cellcolor[HTML]{66ABD4}} \color[HTML]{F1F1F1} 48.0 & {\cellcolor[HTML]{DCEAF6}} \color[HTML]{000000} 23.2 \\
& LGBTQ+ & {\cellcolor[HTML]{F3F8FE}} \color[HTML]{000000} 15.7 & {\cellcolor[HTML]{DFECF7}} \color[HTML]{000000} 22.2 & {\cellcolor[HTML]{CFE1F2}} \color[HTML]{000000} 27.8 & {\cellcolor[HTML]{E5EFF9}} \color[HTML]{000000} 20.3 & {\cellcolor[HTML]{F5F9FE}} \color[HTML]{000000} 15.2 & {\cellcolor[HTML]{084E98}} \color[HTML]{F1F1F1} 72.4 & {\cellcolor[HTML]{BED8EC}} \color[HTML]{000000} 32.4 & {\cellcolor[HTML]{F5F9FE}} \color[HTML]{000000} 15.2 & {\cellcolor[HTML]{F6FAFF}} \color[HTML]{000000} 14.8 & {\cellcolor[HTML]{CBDEF1}} \color[HTML]{000000} 29.1 \\
& Men & {\cellcolor[HTML]{DAE8F6}} \color[HTML]{000000} 24.0 & {\cellcolor[HTML]{9CC9E1}} \color[HTML]{000000} 39.3 & {\cellcolor[HTML]{BCD7EB}} \color[HTML]{000000} 33.0 & {\cellcolor[HTML]{D3E3F3}} \color[HTML]{000000} 26.5 & {\cellcolor[HTML]{D0E2F2}} \color[HTML]{000000} 27.3 & {\cellcolor[HTML]{75B4D8}} \color[HTML]{000000} 45.5 & {\cellcolor[HTML]{6AAED6}} \color[HTML]{F1F1F1} 47.2 & {\cellcolor[HTML]{CEE0F2}} \color[HTML]{000000} 28.2 & {\cellcolor[HTML]{C6DBEF}} \color[HTML]{000000} 31.0 & {\cellcolor[HTML]{99C7E0}} \color[HTML]{000000} 39.9 \\
& Muslims, Arabs & {\cellcolor[HTML]{92C4DE}} \color[HTML]{000000} 40.8 & {\cellcolor[HTML]{A0CBE2}} \color[HTML]{000000} 38.6 & {\cellcolor[HTML]{56A0CE}} \color[HTML]{F1F1F1} 51.5 & {\cellcolor[HTML]{3A8AC2}} \color[HTML]{F1F1F1} 57.3 & {\cellcolor[HTML]{92C4DE}} \color[HTML]{000000} 40.8 & {\cellcolor[HTML]{CEE0F2}} \color[HTML]{000000} 28.0 & {\cellcolor[HTML]{C6DBEF}} \color[HTML]{000000} 30.8 & {\cellcolor[HTML]{083B7C}} \color[HTML]{F1F1F1} 77.0 & {\cellcolor[HTML]{B7D4EA}} \color[HTML]{000000} 34.1 & {\cellcolor[HTML]{C8DCF0}} \color[HTML]{000000} 30.1 \\
& White & {\cellcolor[HTML]{CBDEF1}} \color[HTML]{000000} 29.1 & {\cellcolor[HTML]{A9CFE5}} \color[HTML]{000000} 36.9 & {\cellcolor[HTML]{D0E1F2}} \color[HTML]{000000} 27.6 & {\cellcolor[HTML]{D5E5F4}} \color[HTML]{000000} 25.7 & {\cellcolor[HTML]{AED1E7}} \color[HTML]{000000} 35.7 & {\cellcolor[HTML]{F3F8FE}} \color[HTML]{000000} 15.8 & {\cellcolor[HTML]{D7E6F5}} \color[HTML]{000000} 24.9 & {\cellcolor[HTML]{E7F0FA}} \color[HTML]{000000} 19.8 & {\cellcolor[HTML]{0B559F}} \color[HTML]{F1F1F1} 70.6 & {\cellcolor[HTML]{E8F1FA}} \color[HTML]{000000} 19.3 \\
& Women & {\cellcolor[HTML]{B0D2E7}} \color[HTML]{000000} 35.2 & {\cellcolor[HTML]{64A9D3}} \color[HTML]{F1F1F1} 48.5 & {\cellcolor[HTML]{68ACD5}} \color[HTML]{F1F1F1} 47.7 & {\cellcolor[HTML]{79B5D9}} \color[HTML]{000000} 45.0 & {\cellcolor[HTML]{ABD0E6}} \color[HTML]{000000} 36.2 & {\cellcolor[HTML]{3A8AC2}} \color[HTML]{F1F1F1} 57.3 & {\cellcolor[HTML]{3585BF}} \color[HTML]{F1F1F1} 58.6 & {\cellcolor[HTML]{89BEDC}} \color[HTML]{000000} 42.4 & {\cellcolor[HTML]{94C4DF}} \color[HTML]{000000} 40.7 & {\cellcolor[HTML]{0D57A1}} \color[HTML]{F1F1F1} 70.1 \\[-0.5em]
& & \rot{Asian} & \rot{Black} & \rot{Christians} & \rot{Jews} & \rot{Latinx} & \rot{LGBTQ+} & \rot{Men} & \rot{Muslims, Arabs} & \rot{White} & \rot{Women} \\ [-.5em]
& & \multicolumn{10}{c}{\textbf{Test}} \\
\end{tabular}
\caption{Hate speech classification performance (F1 score) across identity-specific corpora}
\label{fig:identity_generalization}
\end{table}

\begin{figure}[tb]
    \centering
    \resizebox{\columnwidth}{!}{\includegraphics{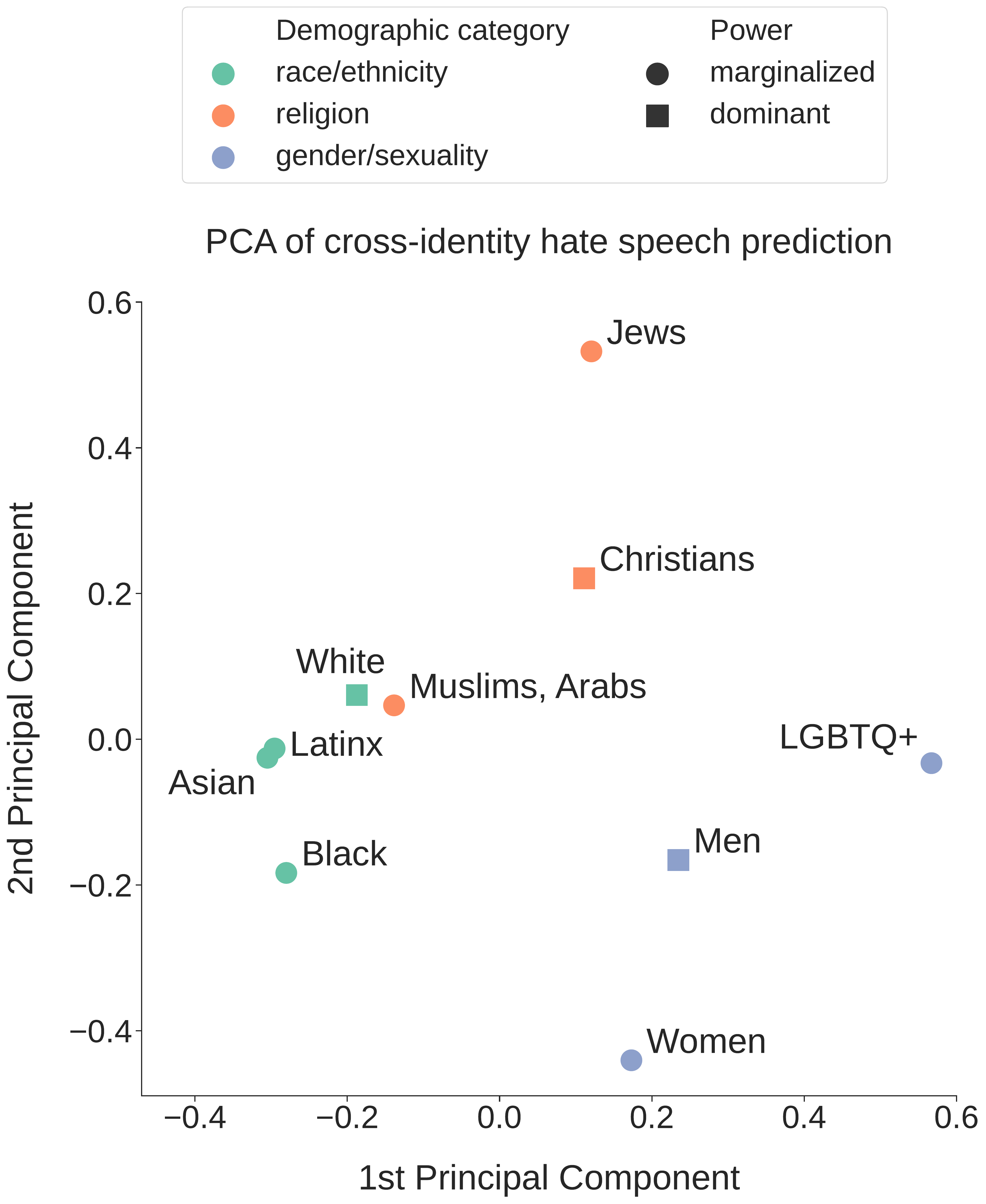}}
    \caption{PCA of cross-identity hate speech classification performance. Hate speech classifiers trained on data targeting identities in the same demographic categories perform most similarly.}
    \label{fig:identity_pca}
\end{figure}

Following \citet{Uyheng2021}, we perform a PCA dimensionality reduction of this generalization performance to 2 factors in order to visualize which target identities exhibit similarities (\autoref{fig:identity_pca}).

Evident from this PCA is a clustering of identity targets by demographic category.
In particular, three clusters are evident: identities that reference religion are in a similar space, while identities that reference race and ethnicity are in a different space, as are terms that reference gender and sexuality.
We look specifically at the effect of these distinctions on hate speech in Section \ref{sec:category}.

Three identities included have relative social power in the European and North American English-speaking contexts from which our datasets were drawn: white people, Christians, and men.
These identities do not form a clear cluster in \autoref{fig:identity_pca}, though they contain factor loadings relatively close to 0 for both factors.
In Section \ref{sec:power}, we investigate how hate speech varies according to the relative social power of the identities targeted.

\section{Variation by Demographic Category}
\label{sec:category}
Poor generalization results across identity targets (\autoref{fig:identity_generalization}) suggest that hate speech varies significantly by the identities it targets.
Our results also suggest that this variation patterns largely by demographic categories such as race/ethnicity, gender/sexuality, and religion (\autoref{fig:identity_pca}).
We hypothesize that if demographic categories are particularly discriminative, hate speech classification performance will drop sharply when attempting to generalize across categories.

To test this, we manually assigned normalized and grouped identities to the categories referenced by the identity.
For example, the identity of `Asian' references race/ethnicity, while `Asian women' references both race/ethnicity and gender/sexuality.
In cases where target groups fit multiple categories (which is not common), we include instances in all corpora they reference.
Though targeted identities sometimes reference categories such as politics, interests, and age, the only categories that met a threshold of 900 hate speech instances uniformly sampled across datasets were race/ethnicity, religion, and gender/sexuality.
Details on corpora constructed by category can be found in \autoref{tab:corpora_stats}.

We then train DistilBERT hate speech classification models on each corpus and test on all others to measure generalization performance in the same way as for identity generalization.
Results can be found in~\autoref{tab:category_generalization}.


\newcolumntype{M}[1]{>{\centering\arraybackslash}m{#1}}

\newcolumntype{R}[2]{%
    >{\adjustbox{angle=#1,lap=\width-(#2)}\bgroup}%
    l%
    <{\egroup}%
}
\newcommand*\rotcategory{\multicolumn{1}{R{-40}{-1.5em}}}

\begin{table}[tb]
    \centering
\setlength\tabcolsep{0pt}
\small
\begin{tabular}{@{\hspace{-3em}}c@{\hspace{-0.8em}}r@{\hspace{0.7em}}@{\rule[-0.5cm]{0pt}{1.2cm}}*{3}{M{1.2cm}}}
\parbox[t]{8mm}{\multirow{8}{*}{\rotatebox[origin=c]{90}{\textbf{Train}}}} & \begin{tabular}{@{}l@{}} Race/ \\ ethnicity \end{tabular} & {\cellcolor[HTML]{084285}} \color[HTML]{F1F1F1} 76.6 & {\cellcolor[HTML]{08519C}} \color[HTML]{F1F1F1} 73.7 & {\cellcolor[HTML]{CADDF0}} \color[HTML]{000000} 41.8 \\
& Religion & {\cellcolor[HTML]{64A9D3}} \color[HTML]{F1F1F1} 56.3 & {\cellcolor[HTML]{083776}} \color[HTML]{F1F1F1} 78.5 & {\cellcolor[HTML]{F7FBFF}} \color[HTML]{000000} 30.2 \\
& \begin{tabular}{@{}l@{}} Gender/ \\ sexuality \end{tabular} & {\cellcolor[HTML]{A4CCE3}} \color[HTML]{000000} 48.1 & {\cellcolor[HTML]{A1CBE2}} \color[HTML]{000000} 48.4 & {\cellcolor[HTML]{1460A8}} \color[HTML]{F1F1F1} 70.9 \\ [-1em]
& & \rotcategory{Race/ethnicity} & \rotcategory{Religion} & \rotcategory{Gender/sexuality} \\[-1.2em]
& & \multicolumn{3}{c}{\hspace{0.5em}\textbf{Test}} \\ [-1em]
\end{tabular}
\caption{Hate speech generalization performance (F1 on hate) by demographic category.
\label{tab:category_generalization}}
\end{table}

Performance drops across identity categories, sometimes falling by almost half of the F1-score.
This suggests that for purposes of automatic classification, hate speech varies significantly by demographic category.
Classifiers generalize particularly poorly from race/ethnicity and religion to gender/sexuality, and less poorly between race/ethnicity and religion.
This may be because of the blurred lines in hate speech targets between racial and religious categories, for example, by conflating Muslims and Arabs or targeting Jews by both religious and racial characteristics.

\section{Variation by Power}
\label{sec:power}
Another significant dimension of variation among targeted identities is relative social power in the societies from which hate speech data has been drawn.
Work on hate speech detection in NLP is often motivated as an effort to fight sexism, racism, homophobia, and other oppressions of marginalized groups, and improve participation of these groups online~\cite{mathew2021hatexplain,Jurgens2019}.
However, this work often frames hate speech as a property of language without considering social context.
Abstracting away from the particulars of targeted identities, datasets often include hate speech directed at any identity group, regardless of the social context of power or marginalization.
Such datasets thus include hate speech directed toward groups with relative social power, such as white people or men in English-speaking European and American contexts.

Calls are growing to consider the role of power and historical oppression in NLP work~\citep{blodgett-etal-2020-language, field-etal-2021-survey}.
Moreover, some theorists of social meaning in language argue that hate speech is fundamentally different when directed at social groups with power \citep{Butler1997,Lakoff2000}.
They note that such speech does not reference the same historical threat of possible violence and recurring oppression as does hate directed toward marginalized groups.
From a lens of social dominance theory~\citep{Sidanius1999}, hate speech serves either to perpetuate or challenge group hierarchies depending on its target.
Activists have called for social media platforms to incorporate this social context by treating hate speech toward marginalized groups as more serious than hate directed toward groups with relative social power \citep{Nurik2019, Dwoskin2020}.

For these theoretical and practical reasons, we consider empirical differences in hate speech based on the social power of targeted identity groups.
Similar to previous experiments, we test the generalization of classifiers across identities with different levels of social power.
We also test for effects on classification performance when removing hate directed toward socially dominant identity groups from hate speech datasets.
If this type of hate is sufficiently different, including it could ``muddy'' the concept we are after and reduce the effectiveness of classifiers in identifying hate speech.
Removing it would more closely match commonly stated motivations of NLP work on hate speech.

\subsection{Generalization}
Just as with demographic categories, we construct separate corpora of hate speech directed at identities with relative social power and identities with relative social marginalization.

We manually label normalized, grouped identity terms with a coarse-grained label as either \textit{dominant}, \textit{marginalized}, or \textit{other}.
This labeling was done by one of the authors familiar with the North American and European English-speaking contexts from which hate speech datasets were drawn.
Identity groups certainly have different social power depending on the setting.
For example, though LGBTQ+ people are generally marginalized, gay men in LGBTQ+ spaces can have higher social power relative to people with more marginalized genders and sexualities~\citep{stulberg2018lgbtq}.
Our goal in annotation was to label identity groups for which there would be broad agreement of enduring dominance or marginalization in North American and European English-speaking societies.
All other cases were marked \textit{other}.
This included political identities such as `Republican' or `liberal', since political power is generally transient in these societies.
Some targeted identities were intersectional, that is, contained multiple identity groups, such as ``white women'' or ``transgender men''.
These cases were taken case-by-case, considering the marginalization of each identity component and marking \textit{other} for many tough cases.
A full list of identities labeled as dominant and marginalized is available in \autoref{tab:marginalized_dominant_identities} in Appendix \ref{sec:appendix}.
Any identities not in these lists were marked \textit{other} by default.

Some datasets all annotators to mark multiple targeted identities.
We marked these instances as directed to \textit{marginalized} groups if there was only \textit{marginalized} or \textit{other} identities targeted.
Instances with both \textit{marginalized} and \textit{dominant} identities targeted were marked as \textit{other}.
Details on corpora constructed by power are in \autoref{tab:corpora_stats}.

As with identities and demographic categories, we evaluated the ability of DistilBERT hate speech classification models to generalize across marginalized and dominant identity targets (\autoref{tab:power_generalization}).



\newcommand*\rotpower{\multicolumn{1}{R{-40}{-1.5em}}}

\begin{table}[tb]
    \centering
\setlength\tabcolsep{0pt}
\small
\begin{tabular}{@{\hspace{-3.5em}}cr@{\hspace{0.7em}}@{\rule[-0.5cm]{0pt}{1.2cm}}*{2}{M{1.2cm}}}
\parbox[t]{5mm}{\multirow{4}{*}{\rotatebox[origin=c]{90}{\textbf{Train}}}} & Dominant & {\cellcolor[HTML]{8DC1DD}} \color[HTML]{000000} 57.9 & {\cellcolor[HTML]{F7FBFF}} \color[HTML]{000000} 42.1 \\
& Marginalized & {\cellcolor[HTML]{69ADD5}} \color[HTML]{F1F1F1} 61.3 & {\cellcolor[HTML]{1561A9}} \color[HTML]{F1F1F1} 72.8 \\ [-1em]
& & \rotpower{Dominant} & \rotpower{Marginalized} \\[-1.2em]
& & \multicolumn{2}{c}{\textbf{Test}} \\ [-1em]
\end{tabular}
\caption{Hate speech generalization performance (F1 on hate) by relative social power.
\label{tab:power_generalization}}
\end{table}

Generalization does not suffer as much across target identities with differences in social power, particularly when trained on the corpus of hate directed at marginalized identities.
This suggests that which target identities have power does not structure variation in hate speech as much as differences in demographic category.

\subsection{Removing hate speech toward socially dominant groups}
\label{sec:power_removal_comparison}
We further evaluate the effect of removing hate speech toward socially dominant groups on classification performance.
We hypothesize that if it is sufficiently different, as some theorists argue, then it may act as noise.
For this experiment, we resample all 7 hate speech datasets listed in \autoref{tab:dataset} separately instead of combining across datasets as in generalization experiments.
This allows us to see trends across even more datasets than we could examine if uniformly sampling from just those with enough to reach a certain threshold.

We resample each dataset to exclude or include hate toward dominant social groups.
All instances are the same between these samples except for instances of hate speech toward dominant social groups and those instances replaced by them.
This allows a comparison across samples of equal size and hate speech ratio.

Removing hate speech toward any set of target identities could improve performance since the remaining instances are more likely to be similar to each other.
For this reason we compare removing hate speech toward dominant groups with removing hate speech toward a set of non-dominant identities.
We select these ``control'' identities to be similar in frequency across datasets to identities labeled as dominant.
Specifically, we match each identity labeled as dominant with the non-dominant identity that has the closest log frequency distribution across datasets (by Euclidean distance).

We perform 5x2-fold cross-validation with a DistilBERT model to estimate performance with and without dominant or control identities.
Parameters are the same as were used with the models built to test generalization, and 10\% of training sets are used as development sets for early stopping.

Two out of the 7 datasets, \citet{elsherief2021latent} and HateXplain, show significant improvement after removing hate speech toward dominant social identities.
However, when removing the control identities, 2 out of the 7 datasets, Civil Comments and HateXplain, also show significant improvements, while the Social Bias Inference Corpus shows a significant decrease in performance.
This does not show convincing evidence that hate speech toward dominant groups is sufficiently different to act as noise for hate speech classification.

\section{Lexical Variation Across Target Identities}
To explore how hate speech varies by target identity, we examine the words most strongly associated with each target identity and grouping of identities.
We use the Sparse Additive Generative Model~\citep[SAGE;][]{Eisenstein2011} to find words that are most representative of each hate speech corpus.
SAGE finds representative words by learning a generative model that contrasts terms in documents in a section of a corpus with a background frequency distribution over the whole corpus.
We run SAGE over 3 separate corpora: one where each section is an identity-specific split, another with category splits, and another with splits by relative social power.
We run SAGE with a vocabulary size of the most frequent 3000 words and a smoothing rate of 50.
Larger vocabulary sizes and lower smoothing included less informative, specialty words that did not occur frequently in the corpus.
The 10 most representative terms for each of these splits are shown in \autoref{tab:top_terms}.

\begin{table*}[tb]
\begin{tabular}{ll}
\toprule
\textit{Identity} & Top terms \\
\midrule
Asian & chinese, china, asian, ching, chong, asians, japanese, chinaman, ch*nk, japan \\
Black & n*ggas, black, n*gga, n*gger, africa, blm, negro, ethiopian, blacks, african \\
Christians & priest, catholic, jesus, priests, bible, christians, christianity, christian, church \\
Jews & jewish, jews, holocaust, jew, israel, hitler, gas, oven, zionist, k*ke \\
Latinx & latinos, latino, mexico, mexican, mexicans, beaner, sp*c, latin, hispanic, beaners \\
LGBTQ+ & transgender, transgendered, transgenders, bisexual, queers, bathroom, f*g, gay \\
Men & divorce, dudes, men, male, negative, movies, man, priests, soy, dad \\
Muslims, Arabs & islam, muslim, islamic, muslims, isis, terrorist, terrorists, iran, bomb, radical \\
White & redneck, white, supremacist, supremacy, mudshark, trash, fascist, shootings \\
Women & hoes, sexist, woman, hoe, feminist, women, feminists, feminism, slut, bitches \\
\midrule

\textit{Category} & \\
\midrule
Gender/sexuality & hoes, dyke, transgender, f*ggot, f*g, sexist, sexual, lesbian, hoe, dykes \\
Race/ethnicity & chinese, black, blacks, asian, asians, mexicans, whites, africa, supremacist \\
Religion & catholic, priest, christians, christian, christianity, religion, church, jesus, koran \\
\midrule

\textit{Power} &  \\
\midrule
Dominant & priest, catholic, priests, jesus, catholics, virgin, church, devil, dress \\
Marginalized & muslim, muslims, she, islam, her, woman, n*gger, black, jews, women \\
\bottomrule
\end{tabular}
\caption{Most representative terms (lowercased) in corpora divided by different target identity sets from SAGE.}
\label{tab:top_terms}
\end{table*}

Identity terms, many of them derogatory, form the bulk of these representative words.
This provides more evidence for the centrality of identities to hate speech~\citep{Uyheng2021}.
Some representative words relate to identity-specific histories of oppression.
For example, `oven' and `gas' are representative terms of antisemitic hate speech.
Identity-specific stereotypes are also visible: `terrorist' and `bomb' are top terms in hate speech against Muslims and Arabs.
Current culture wars issues are also relevant.
For example, transphobic attitudes around bathrooms are reflected in the top terms in hate speech targeting LGBTQ+ people.
`BLM', for the Black Lives Matter movement, is a top term associated with anti-Black hate speech.

The difficulty in a binary distinction of dominance and marginalization can be seen through the most representative words in hate directed toward groups with high relative social power.
As a marker of Christianity, `Catholic', for example, could be seen as dominant in European and American contexts where Christianity has historically been a religion with relative social and cultural prominence.
However, some white nationalist groups such as the Ku Klux Klan have targeted Catholics as outside idealized Christian Protestantism~\cite{Burris2000,Berlet2006}.
`Redneck' and `trash' are top terms in hate targeting white people, and `virgin', a top term in hate targeting dominant groups, is used in jokes stereotyping incest.
Such terms target poor white people based mainly on class.
Also in the top terms against white people is `mudshark', a derogatory term targeting white women who have relationships with Black men.
These terms target groups that are marginalized within broadly dominant groups: white women, poor white people, and Catholics.
Such examples show how social power is relative, complex, and intersectional.
They also evidence a tendency for hate speech to target marginalized groups, even within groups that have higher relative social power.

\section{Discussion}
Our results demonstrate that hate speech varies considerably according to which identities are targeted.
We show evidence that classifiers trained on hate toward one target identity generalize poorly to other target identities, especially across demographic categories such as race/ethnicity, religion and gender/sexuality.

These results suggest that the designers of hate speech classifiers pay attention to the distribution of targeted identities in training data.
Many commonly used hate speech datasets do not specify this information.
If the distribution skews toward a particular identity group (such as anti-Black racism), then using such a classifier on data that has a different distribution (e.g., mostly antisemitic) would likely give poor performance.
More generally, these results suggest a value in treating hate speech as a social and linguistic category with lots of internal variance.
This variance depends in part on the social context around targeted identities.

Classifiers trained on hate speech toward dominant or marginalized groups suffered somewhat when tested on the opposite group.
However, we did not find evidence that removing hate speech toward dominant groups clarifies the hate speech signal enough to consistently increase performance beyond what might be expected by removing a random set of targeted identities.
This suggests that differences based on the social context of power do not affect the language of hate speech enough to be easily detectable by machine learning classifiers.
Differences in severity between hate speech targeting socially marginalized or powerful groups is more likely a matter of interpretation by those with social knowledge of power in a particular society.

\section{Conclusion}
We present a meta-analysis of hate speech datasets annotated for identity group targets.
This analysis shows that hate speech differs significantly by target identity, as classifiers trained on hate speech toward one identity do not generalize well to other identities.
We then examine what factors of social context structure this variation by target identity.
We find evidence for hate speech varying substantially by demographic category, and less so by the relative social power of targeted identities.

These results reinforce the importance of variation by social context within hate speech and suggest that researchers pay attention to variation by target identity.
Future work may address improving generalization across target identities by strategically sampling training data or incorporating multiple identity-specific classifiers.
Similar analyses may also be conducted on multilingual hate speech datasets in future work.

\section{Limitations and Ethics}
As a meta-analysis of existing datasets, this study is limited by the availability of hate speech data labeled with target identity.
Performance estimates with and without hate speech toward dominant groups would be more reliable with more labeled hate speech toward socially dominant groups.
The scarcity of hate speech against socially dominant groups is not coincidental: this speech is less prototypically considered hate speech than that against marginalized groups.
This can be seen in the dataset from \citet{Kennedy2020}, for example, where annotators rate the average severity of hate against dominant groups as less than the average severity of hate against marginalized groups.

Another limitation is that datasets each have their own definitions of hate speech and associated annotation criteria, which may vary considerably.
We attempted to mitigate the effects of any one dataset's definition with uniform sampling (see Section \ref{sec:sampling}).
Since we take these annotations as representative of hate speech, it is necessary to be mindful that we are not capturing any true sense of ``hate speech'', but simply what annotators have identified as hate speech.
However, we wished to investigate the role of target identity in existing hate speech classification approaches, for which existing datasets and their associated definitions are most relevant.

These datasets are only in English and largely reflect European and American societies.
Our findings are specific to this context. 
Experiments on multilingual datasets may reveal other trends and reflect different social associations around identity terms, which are culturally specific.

When sampling identity-based corpora from datasets, we attempted to control for the idiosyncrasies of any particular dataset.
However, the sizes of the resulting identity-specific corpora vary depending on how much hate speech directed toward them occurs across datasets.
This could influence our generalization experiments.
Classifiers trained on identities with small corpora still perform well on test sets of identities with the same demographic category, the general trend we report.
As seen in \autoref{fig:identity_pca}, identities with lots of data sometimes exhibit behavior similar to identities with not as much data.
These factors lead us to doubt that corpus size has a large impact on generalization results.

Care must always be taken to specify that differences based on identity, in this case hate speech directed toward identities, are due to social, not biological, factors \citep{Hanna2020,Lu2022}.
We attempt to be clear that these differences are the result of social context.

\section*{Acknowledgements}
This work was supported in part by the Collaboratory Against Hate: Research and Action Center at Carnegie Mellon University and the University of Pittsburgh. 
The Center for Informed Democracy and Social Cybersecurity at Carnegie Mellon University also provided support.
We thank the researchers who made their annotated hate speech data publicly available, which enabled this meta-analysis.

\bibliography{anthology,custom}

\begin{thebibliography}{52}
\expandafter\ifx\csname natexlab\endcsname\relax\def\natexlab#1{#1}\fi

\bibitem[{Basile et~al.(2019)Basile, Bosco, Fersini, Nozza, Patti, Rangel,
  Rosso, and Sanguinetti}]{Basile2019}
Valerio Basile, Cristina Bosco, Elisabetta Fersini, Debora Nozza, Viviana
  Patti, Francisco Rangel, Paolo Rosso, and Manuela Sanguinetti. 2019.
\newblock \href {http://evalita.org} {{SemEval-2019 Task 5: Multilingual
  Detection of Hate Speech Against Immigrants and Women in Twitter}}.
\newblock In \emph{Proceedings of the 13th International Workshop on Semantic
  Evaluation (SemEval-2019)}, pages 54--63.

\bibitem[{Berlet and Vysotsky(2006)}]{Berlet2006}
Chip Berlet and Stanislav Vysotsky. 2006.
\newblock {Overview of U.S. white supremacist groups}.
\newblock \emph{Journal of Political and Military Sociology}, 34:11--48.

\bibitem[{Blodgett et~al.(2020)Blodgett, Barocas, Daum{\'e}~III, and
  Wallach}]{blodgett-etal-2020-language}
Su~Lin Blodgett, Solon Barocas, Hal Daum{\'e}~III, and Hanna Wallach. 2020.
\newblock \href {https://doi.org/10.18653/v1/2020.acl-main.485} {Language
  (technology) is power: A critical survey of {``}bias{''} in {NLP}}.
\newblock In \emph{Proceedings of the 58th Annual Meeting of the Association
  for Computational Linguistics}, pages 5454--5476, Online. Association for
  Computational Linguistics.

\bibitem[{Borkan et~al.(2019)Borkan, Dixon, Sorensen, Thain, and
  Vasserman}]{Borkan2019}
Daniel Borkan, Lucas Dixon, Jeffrey Sorensen, Nithum Thain, and Lucy Vasserman.
  2019.
\newblock \href {https://doi.org/10.1145/3308560.3317593} {Nuanced metrics for
  measuring unintended bias with real data for text classification}.
\newblock In \emph{Companion Proceedings of the 2019 World Wide Web
  Conference}, pages 491--500. Association for Computing Machinery.

\bibitem[{Brown(2017)}]{Brown2017part2}
Alexander Brown. 2017.
\newblock \href {https://doi.org/10.1007/s10982-017-9300-x} {What is hate
  speech? part 2: Family resemblances}.
\newblock \emph{Law and Philosophy}, 36:561--613.

\bibitem[{Burris et~al.(2000)Burris, Smith, and Strahm}]{Burris2000}
Val Burris, Emery Smith, and Ann Strahm. 2000.
\newblock {White supremacist networks on the Internet}.
\newblock \emph{Source: Sociological Focus}, 33:215--235.

\bibitem[{Butler(1997)}]{Butler1997}
Judith Butler. 1997.
\newblock \href {https://doi.org/10.4324/9780203948682} {\emph{Excitable
  Speech}}, 1st edition.
\newblock Routledge.

\bibitem[{Devlin et~al.(2019)Devlin, Chang, Lee, and
  Toutanova}]{devlin-etal-2019-bert}
Jacob Devlin, Ming-Wei Chang, Kenton Lee, and Kristina Toutanova. 2019.
\newblock \href {https://doi.org/10.18653/v1/N19-1423} {{BERT}: Pre-training of
  deep bidirectional transformers for language understanding}.
\newblock In \emph{Proceedings of the 2019 Conference of the North {A}merican
  Chapter of the Association for Computational Linguistics: Human Language
  Technologies, Volume 1 (Long and Short Papers)}, pages 4171--4186,
  Minneapolis, Minnesota. Association for Computational Linguistics.

\bibitem[{Dixon et~al.(2017)Dixon, Li, Sorensen, Thain, and
  Vasserman}]{Dixon2017}
Lucas Dixon, John Li, Jeffrey Sorensen, Nithum Thain, and Lucy Vasserman. 2017.
\newblock \href
  {http://www.aies-conference.com/wp-content/papers/main/AIES_2018_paper_9.pdf}
  {Measuring and mitigating unintended bias in text classification}.
\newblock In \emph{AAAI/ACM Conference on Artificial Intelligence, Ethics, and
  Society (AIES)}.

\bibitem[{Dwoskin et~al.(2020)Dwoskin, Tiku, and Kelly}]{Dwoskin2020}
Elizabeth Dwoskin, Nitasha Tiku, and Heather Kelly. 2020.
\newblock \href
  {https://www.washingtonpost.com/technology/2020/12/03/facebook-hate-speech/}
  {Facebook to start policing anti-black hate speech more aggressively than
  anti-white comments, documents show}.
\newblock \emph{The Washington Post}.

\bibitem[{Eisenstein et~al.(2011)Eisenstein, Ahmed, and Xing}]{Eisenstein2011}
Jacob Eisenstein, Amr Ahmed, and Eric~P. Xing. 2011.
\newblock Sparse additive generative models of text.
\newblock In \emph{Proceedings of the 28th International Conference on Machine
  Learning}, pages 1041--1048.

\bibitem[{ElSherief et~al.(2018{\natexlab{a}})ElSherief, Kulkarni, Nguyen,
  Wang, and Belding}]{elsherief2018hate}
Mai ElSherief, Vivek Kulkarni, Dana Nguyen, William~Yang Wang, and Elizabeth
  Belding. 2018{\natexlab{a}}.
\newblock Hate lingo: A target-based linguistic analysis of hate speech in
  social media.
\newblock In \emph{Proceedings of the International AAAI Conference on Web and
  Social Media}, volume~12.

\bibitem[{ElSherief et~al.(2018{\natexlab{b}})ElSherief, Nilizadeh, Nguyen,
  Vigna, and Belding}]{ElSherief2018}
Mai ElSherief, Shirin Nilizadeh, Dana Nguyen, Giovanni Vigna, and Elizabeth
  Belding. 2018{\natexlab{b}}.
\newblock Peer to peer hate: Hate speech instigators and their targets.
\newblock In \emph{Proceedings of the Twelfth International AAAI Conference on
  Web and Social Media (ICWSM 2018)}, pages 52--61.

\bibitem[{ElSherief et~al.(2021)ElSherief, Ziems, Muchlinski, Anupindi,
  Seybolt, De~Choudhury, and Yang}]{elsherief2021latent}
Mai ElSherief, Caleb Ziems, David Muchlinski, Vaishnavi Anupindi, Jordyn
  Seybolt, Munmun De~Choudhury, and Diyi Yang. 2021.
\newblock Latent hatred: A benchmark for understanding implicit hate speech.
\newblock In \emph{Proceedings of the 2021 Conference on Empirical Methods in
  Natural Language Processing}, pages 345--363.

\bibitem[{Field et~al.(2021)Field, Blodgett, Waseem, and
  Tsvetkov}]{field-etal-2021-survey}
Anjalie Field, Su~Lin Blodgett, Zeerak Waseem, and Yulia Tsvetkov. 2021.
\newblock \href {https://doi.org/10.18653/v1/2021.acl-long.149} {A survey of
  race, racism, and anti-racism in {NLP}}.
\newblock In \emph{Proceedings of the 59th Annual Meeting of the Association
  for Computational Linguistics and the 11th International Joint Conference on
  Natural Language Processing (Volume 1: Long Papers)}, pages 1905--1925,
  Online. Association for Computational Linguistics.

\bibitem[{Fortuna et~al.(2020)Fortuna, Soler, and
  Wanner}]{fortuna-etal-2020-toxic}
Paula Fortuna, Juan Soler, and Leo Wanner. 2020.
\newblock \href {https://aclanthology.org/2020.lrec-1.838} {Toxic, hateful,
  offensive or abusive? what are we really classifying? an empirical analysis
  of hate speech datasets}.
\newblock In \emph{Proceedings of the 12th Language Resources and Evaluation
  Conference}, pages 6786--6794, Marseille, France. European Language Resources
  Association.

\bibitem[{Gröndahl et~al.(2018)Gröndahl, Pajola, Juuti, Conti, and
  Asokan}]{Groendahl2018}
Tommi Gröndahl, Luca Pajola, Mika Juuti, Mauro Conti, and N.~Asokan. 2018.
\newblock \href {https://doi.org/10.1145/3270101.3270103} {All you need is
  “love”: Evading hate speech detection}.
\newblock In \emph{Proceedings of the 11th ACM Workshop on Artificial
  Intelligence and Security (AISec ’18)}, pages 2--12. Association for
  Computing Machinery.

\bibitem[{Hanna et~al.(2020)Hanna, Denton, Smart, and Smith-Loud}]{Hanna2020}
Alex Hanna, Emily Denton, Andrew Smart, and Jamila Smith-Loud. 2020.
\newblock \href {https://doi.org/10.1145/3351095.3372826} {Towards a critical
  race methodology in algorithmic fairness}.
\newblock In \emph{FAT* 2020 - Proceedings of the 2020 Conference on Fairness,
  Accountability, and Transparency}, pages 501--512.

\bibitem[{Jurgens et~al.(2019)Jurgens, Chandrasekharan, and
  Hemphill}]{Jurgens2019}
David Jurgens, Eshwar Chandrasekharan, and Libby Hemphill. 2019.
\newblock {A just and comprehensive strategy for using NLP to address online
  abuse}.
\newblock In \emph{Proceedings of the 57th Annual Meeting of the Association
  for Computational Linguistics}, pages 3658--3666. Association for
  Computational Lingustics.

\bibitem[{Kennedy et~al.(2020)Kennedy, Bacon, Sahn, and von
  Vacano}]{Kennedy2020}
Chris~J. Kennedy, Geoff Bacon, Alexander Sahn, and Claudia von Vacano. 2020.
\newblock \href {http://arxiv.org/abs/2009.10277} {{Constructing interval
  variables via faceted Rasch measurement and multitask deep learning: a hate
  speech application}}.

\bibitem[{Kwok and Wang(2013)}]{Kwok2013}
Irene Kwok and Yuzhou Wang. 2013.
\newblock \href
  {http://www.google.com/url?sa=t&amp;rct=j&amp;q=&amp;esrc=s&amp;source=web&amp;cd=1&amp;ved=0CC0QFjAA&amp;url=http://www.aaai.org/ocs/index.php/AAAI/AAAI13/paper/download/6419/6821&amp;ei=e7hJUq2EAtKq4AOB04HoDg&amp;usg=AFQjCNEi9mX0w71lUCo8tdxTnQJkR74MLg&am}
  {Locate the hate: Detecting tweets against blacks}.
\newblock In \emph{Twenty-Seventh AAAI Conference on Artificial Intelligence},
  pages 1621--1622.

\bibitem[{Lakoff(2000)}]{Lakoff2000}
Robin~Tolmach Lakoff. 2000.
\newblock \href {https://doi.org/10.1525/j.ctt1pp38b.7} {\emph{The Language
  War}}.
\newblock University of California Press.

\bibitem[{Lu et~al.(2022)Lu, Kay, and McKee}]{Lu2022}
Christina Lu, Jackie Kay, and Kevin~R. McKee. 2022.
\newblock \href {https://doi.org/10.1145/3531146.3533161} {Subverting machines,
  fluctuating identities: Re-learning human categorization}.
\newblock In \emph{FAccT '22: 2022 ACM Conference on Fairness, Accountability,
  and Transparency}, pages 1005--1014.

\bibitem[{Marsters(2019)}]{Marsters2019}
Alexandria Marsters. 2019.
\newblock \emph{When hate speech leads to hateful actions: A corpus and
  discourse analytic approach to linguistic threat assessment of hate speech}.
\newblock Ph.D. thesis, Georgetown University.

\bibitem[{Mathew et~al.(2021)Mathew, Saha, Yimam, Biemann, Goyal, and
  Mukherjee}]{mathew2021hatexplain}
Binny Mathew, Punyajoy Saha, Seid~Muhie Yimam, Chris Biemann, Pawan Goyal, and
  Animesh Mukherjee. 2021.
\newblock Hatexplain: A benchmark dataset for explainable hate speech
  detection.
\newblock In \emph{Proceedings of the AAAI Conference on Artificial
  Intelligence}, volume~35, pages 14867--14875.

\bibitem[{Mondal et~al.(2017)Mondal, Silva, and Benevenuto}]{Mondal2017}
Mainack Mondal, Leandro~Araújo Silva, and Fabrício Benevenuto. 2017.
\newblock \href {https://doi.org/10.1145/3078714.3078723} {A measurement study
  of hate speech in social media}.
\newblock In \emph{HT 2017 - Proceedings of the 28th ACM Conference on
  Hypertext and Social Media}, pages 85--94. Association for Computing
  Machinery.

\bibitem[{Mozafari et~al.(2019)Mozafari, Farahbakhsh, and
  Crespi}]{Mozafari2019}
Marzieh Mozafari, Reza Farahbakhsh, and Noël Crespi. 2019.
\newblock {A BERT-Based Transfer Learning Approach for Hate Speech Detection in
  Online Social Media}.
\newblock In \emph{International Conference on Complex Networks and Their
  Applications.}, pages 928--940.

\bibitem[{Nurik(2019)}]{Nurik2019}
Chloé Nurik. 2019.
\newblock {"Men Are Scum": Self-Regulation, Hate Speech, and Gender-Based
  Censorship on Facebook}.
\newblock \emph{International Journal of Communication}, 13:2878--2898.

\bibitem[{Ousidhoum et~al.(2019)Ousidhoum, Lin, Zhang, Song, and
  Yeung}]{ousidhoum_multilingual_2019}
Nedjma Ousidhoum, Zizheng Lin, Hongming Zhang, Yangqiu Song, and Dit-Yan Yeung.
  2019.
\newblock Multilingual and {Multi}-{Aspect} {Hate} {Speech} {Analysis}.
\newblock In \emph{Proceedings of the 2019 Conference on Empirical Methods in
  Natural Language Processing and the 9th International Joint Conference on
  Natural Language Processing}, pages 4675--4684.

\bibitem[{Pamungkas et~al.(2020)Pamungkas, Basile, and Patti}]{Pamungkas2020}
Endang~Wahyu Pamungkas, Valerio Basile, and Viviana Patti. 2020.
\newblock \href {https://doi.org/10.1016/j.ipm.2020.102360} {Misogyny detection
  in twitter: a multilingual and cross-domain study}.
\newblock \emph{Information Processing and Management}, 57.

\bibitem[{Pohjonen and Udupa(2017)}]{Pohjonen2017}
Matti Pohjonen and Sahana Udupa. 2017.
\newblock \href {https://ijoc.org/index.php/ijoc/article/download/5843/1965}
  {Extreme speech online: An anthropological critique of hate speech debates}.
\newblock \emph{International Journal of Communication}, 11:1173--1191.

\bibitem[{Poletto et~al.(2021)Poletto, Basile, Sanguinetti, Bosco, and
  Patti}]{Poletto2021}
Fabio Poletto, Valerio Basile, Manuela Sanguinetti, Cristina Bosco, and Viviana
  Patti. 2021.
\newblock \href {https://doi.org/10.1007/s10579-020-09502-8} {Resources and
  benchmark corpora for hate speech detection: a systematic review}.
\newblock In \emph{Language Resources and Evaluation}, volume~55, pages
  477--523. Springer Science and Business Media B.V.

\bibitem[{Qian et~al.(2018)Qian, Elsherief, Belding, and
  Wang}]{qian_leveraging_2018}
Jing Qian, Mai Elsherief, Elizabeth~M Belding, and William~Yang Wang. 2018.
\newblock Leveraging {Intra}-{User} and {Inter}-{User} {Representation}
  {Learning} for {Automated} {Hate} {Speech} {Detection}.
\newblock In \emph{Proceedings of {NAACL}-{HLT} 2018}, pages 118--123.

\bibitem[{Rieger et~al.(2021)Rieger, Kümpel, Wich, Kiening, and
  Groh}]{rieger_assessing_2021}
Diana Rieger, Anna~Sophie Kümpel, Maximilian Wich, Toni Kiening, and Georg
  Groh. 2021.
\newblock \href {https://doi.org/10.1177/20563051211052906} {Assessing the
  {Extent} and {Types} of {Hate} {Speech} in {Fringe} {Communities}: {A} {Case}
  {Study} of {Alt}-{Right} {Communities} on 8chan, 4chan, and {Reddit}}.
\newblock \emph{Social Media and Society}, 7(4).

\bibitem[{Salminen et~al.(2018)Salminen, Almerekhi, Milenkovi{\'c}, Jung, An,
  Kwak, and Jansen}]{salminen2018anatomy}
Joni Salminen, Hind Almerekhi, Milica Milenkovi{\'c}, Soon-gyo Jung, Jisun An,
  Haewoon Kwak, and Bernard~J Jansen. 2018.
\newblock Anatomy of online hate: developing a taxonomy and machine learning
  models for identifying and classifying hate in online news media.
\newblock In \emph{Twelfth International AAAI Conference on Web and Social
  Media}, pages 330--339.

\bibitem[{Samghabadi et~al.(2020)Samghabadi, Patwa, Pykl, Mukherjee, Das, and
  Solorio}]{Samghabadi2020}
Niloofar~Safi Samghabadi, Parth Patwa, Srinivas Pykl, Prerana Mukherjee,
  Amitava Das, and Thamar Solorio. 2020.
\newblock Aggression and misogyny detection using bert: A multi-task approach.
\newblock In \emph{Proceedings of the Second Workshop on Trolling, Aggression
  and Cyberbullying}, pages 11--16.

\bibitem[{Sanguinetti et~al.(2018)Sanguinetti, Poletto, Bosco, Patti, and
  Stranisci}]{Sanguinetti2018}
Manuela Sanguinetti, Fabio Poletto, Cristina Bosco, Viviana Patti, and Marco
  Stranisci. 2018.
\newblock {An Italian Twitter Corpus of Hate Speech against Immigrants}.
\newblock In \emph{Proceedings of the Eleventh International Conference on
  Language Resources and Evaluation (LREC’18)}, pages 2798--2895.

\bibitem[{Sanh et~al.(2019)Sanh, Debut, Chaumond, and Wolf}]{Sanh2019}
Victor Sanh, Lysandre Debut, Julien Chaumond, and Thomas Wolf. 2019.
\newblock {DistilBERT, a distilled version of BERT: smaller, faster, cheaper
  and lighter}.
\newblock In \emph{5th Workshop on Energy Efficient Machine Learning and
  Cognitive Computing}.

\bibitem[{Sap et~al.(2020)Sap, Gabriel, Qin, Jurafsky, Smith, and
  Choi}]{Sap2020}
Maarten Sap, Saadia Gabriel, Lianhui Qin, Dan Jurafsky, Noah~A. Smith, and
  Yejin Choi. 2020.
\newblock \href {https://doi.org/10.18653/v1/2020.acl-main.486} {Social bias
  frames: Reasoning about social and power implications of language}.
\newblock In \emph{Proceedings of the 58th Annual Meeting of the Association
  for Computational Linguistics}, pages 5477--5490.

\bibitem[{Sap et~al.(2022)Sap, Swayamdipta, Vianna, Zhou, Choi, and
  Smith}]{sap2022annotators}
Maarten Sap, Swabha Swayamdipta, Laura Vianna, Xuhui Zhou, Yejin Choi, and
  Noah~A Smith. 2022.
\newblock Annotators with attitudes: How annotator beliefs and identities bias
  toxic language detection.
\newblock In \emph{Proceedings of the 2022 Conference of the North American
  Chapter of the Association for Computational Linguistics: Human Language
  Technologies}.

\bibitem[{Sidanius and Pratto(1999)}]{Sidanius1999}
Jim Sidanius and Felicia Pratto. 1999.
\newblock \href {https://doi.org/10.1017/cbo9781139175043.002} {\emph{Social
  Dominance}}.
\newblock Cambridge University Press.

\bibitem[{Silva et~al.(2016)Silva, Mondal, Correa, Benevenuto, and
  Weber}]{Silva2016}
Leandro Silva, Mainack Mondal, Denzil Correa, Fabricio Benevenuto, and Ingmar
  Weber. 2016.
\newblock \href {http://arxiv.org/abs/1603.07709} {Analyzing the targets of
  hate in online social media}.
\newblock In \emph{Proceedings of the Tenth International AAAI Conference on
  Web and Social Media (ICWSM 2016)}, pages 687--690.

\bibitem[{Stulberg(2018)}]{stulberg2018lgbtq}
Lisa~M Stulberg. 2018.
\newblock \emph{LGBTQ social movements}.
\newblock John Wiley \& Sons.

\bibitem[{Swamy et~al.(2019)Swamy, Jamatia, and Gambäck}]{Swamy2019}
Steve~Durairaj Swamy, Anupam Jamatia, and Björn Gambäck. 2019.
\newblock Studying generalisability across abusive language detection datasets.
\newblock In \emph{Proceedings of the 23rd Conference on Computational Natural
  Language Learning}, pages 940--950. Association for Computational
  Linguistics.

\bibitem[{Unsvåg and Gambäck(2018)}]{unsvag_effects_2018}
Elise~Fehn Unsvåg and Björn Gambäck. 2018.
\newblock The {Effects} of {User} {Features} on {Twitter} {Hate} {Speech}
  {Detection}.
\newblock In \emph{Proceedings of the Second Workshop on Abusive Language
  Online (ALW2)}, pages 75--85.

\bibitem[{Uyheng and Carley(2021)}]{Uyheng2021}
Joshua Uyheng and Kathleen~M. Carley. 2021.
\newblock \href {https://doi.org/10.1007/978-3-030-80387-2_12} {An
  identity-based framework for generalizable hate speech detection}.
\newblock In \emph{International Conference on Social Computing,
  Behavioral-Cultural Modeling and Prediction and Behavior Representation in
  Modeling and Simulation}, pages 121--130.

\bibitem[{Vidgen and Derczynski(2020)}]{Vidgen2020}
Bertie Vidgen and Leon Derczynski. 2020.
\newblock \href {https://doi.org/10.1371/journal.pone.0243300} {Directions in
  abusive language training data, a systematic review: Garbage in, garbage
  out}.
\newblock \emph{PLoS ONE}, 15.

\bibitem[{Vidgen et~al.(2021)Vidgen, Nguyen, Margetts, Rossini, and
  Tromble}]{vidgen-etal-2021-introducing}
Bertie Vidgen, Dong Nguyen, Helen Margetts, Patricia Rossini, and Rebekah
  Tromble. 2021.
\newblock \href {https://doi.org/10.18653/v1/2021.naacl-main.182} {Introducing
  {CAD}: the contextual abuse dataset}.
\newblock In \emph{Proceedings of the 2021 Conference of the North American
  Chapter of the Association for Computational Linguistics: Human Language
  Technologies}, pages 2289--2303, Online. Association for Computational
  Linguistics.

\bibitem[{Waseem(2016)}]{Waseem2016}
Zeerak Waseem. 2016.
\newblock \href {https://doi.org/10.18653/v1/W16-5618} {{Are You a Racist or Am
  I Seeing Things? Annotator Influence on Hate Speech Detection on Twitter}}.
\newblock In \emph{Proceedings of the First Workshop on NLP and Computational
  Social Science}, pages 138--142.

\bibitem[{Waseem and Hovy(2016)}]{Waseem_hovy2016}
Zeerak Waseem and Dirk Hovy. 2016.
\newblock \href {https://doi.org/10.18653/v1/N16-2013} {Hateful {Symbols} or
  {Hateful} {People}? {Predictive} {Features} for {Hate} {Speech} {Detection}
  on {Twitter}}.
\newblock In \emph{Proceedings of the {NAACL}-{HLT} 2016}, pages 88--93.

\bibitem[{Wiegand et~al.(2021)Wiegand, Ruppenhofer, and
  Eder}]{wiegand-etal-2021-implicitly-abusive}
Michael Wiegand, Josef Ruppenhofer, and Elisabeth Eder. 2021.
\newblock \href {https://doi.org/10.18653/v1/2021.naacl-main.48} {Implicitly
  abusive language {--} what does it actually look like and why are we not
  getting there?}
\newblock In \emph{Proceedings of the 2021 Conference of the North American
  Chapter of the Association for Computational Linguistics: Human Language
  Technologies}, pages 576--587, Online. Association for Computational
  Linguistics.

\bibitem[{Yin and Zubiaga(2021)}]{Yin2021}
Wenjie Yin and Arkaitz Zubiaga. 2021.
\newblock \href {https://doi.org/10.7717/PEERJ-CS.598} {Towards generalisable
  hate speech detection: a review on obstacles and solutions}.
\newblock \emph{PeerJ Computer Science}, 7:1--38.

\end{thebibliography}
\bibliographystyle{acl_natbib}

\appendix
\section{Appendix}
\label{sec:appendix}
\begin{table*}[tb]
\centering
\begin{tabular}{p{2cm}p{13cm}}
\hline
\textbf{Marginalized} & women, people with mental disabilities, black people, gay men, transgender people, muslims, jewish people, gay people, sexual and gender minorities, feminists, chinese women, people with autism, lgbtqa community, people from china, illegal immigrants, people from pakistan, working class people, elderly people, non-white people, people from mexico, people from india, people with aspergers, people with mental health issues, people with disabilities, romani people, ethnic minorities, immigrants, minorities, jews, blacks, black folks, illegals, people of color, non-whites, islamic people, gays, mexicans, illegal aliens, arabs, africans, refugees, indians, hispanics, black men, arabians, hindus, black lives matter, iranians, mexican, latino folks, asian folks, foreigners, jewish folks, muslim folks, latino/latina folks, physically disabled folks, mentally disabled folks, lesbian women, folks with mental illness/disorder, holocaust victims, native american/first nation folks, trans women, arabic folks, folks with physical illness/disorder, overweight/fat folks, trans men, rape victims, bisexual women, children, poor folks, african folks, ethiopians, bisexual men, sexual assault victims, harassment victims, africa, old folks, orphans, mexican folks, indian folks, child rape victims, ethiopian folks, child sexual assault victims, young children, ethiopian, genocide victims, pregnant folks, ethiopia, pedophilia victims, kids, japanese, chinese folks, holocaust survivors, asian, black, latinx, middle eastern, native american, pacific islander, hindu, jewish, muslim, immigrant, migrant worker, undocumented, non\_binary, transgender\_men, transgender\_unspecified, transgender\_women, bisexual, gay, lesbian, seniors, disability\_physical, disability\_cognitive, disability\_neurological, disability\_visually\_impaired, disability\_hearing\_impaired, disability\_unspecific, disability\_other, disability, xenophobia, islam, jews/judaism, special\_needs, african\_descent, indian/hindu, asians, asian people, muslims and arabic/middle eastern people, lgbtq+ people, victims of violence, non-binary people, older people, bisexual people, chinese people, arabic/middle eastern people, african people, indian people, ethiopian people, japanese people, mexican people, transgender men, undocumented immigrants, latinx people, native american people, people with physical disabilities, transgender women, buddhists, indigenous people, gay or lesbian people, gay and lesbian people \\ \hline 
\textbf{Dominant} & involuntary celibates, white people, police officers, people from america, men, christians, rich people, white men, whites, white folks, conservative males, white conservatives, white liberals, americans, white nationalists, male conservatives, cops, police, white, conservative men, christian folks, christian, straight, middle\_aged, law enforcement, wealthy people, corporations, military, armed forces, straight people, middle-aged people \\ \hline 
\textbf{Other} & left-wing people, moderators, liberals, communists, left-wing people (social justice), non-gender dysphoric transgender people, right-wing people, democrats, activists (anti-fascist), donald trump supporters, republicans, conservatives, gamers, activists (animal rights), people with drug problems, fans of anthropomorphic animals (``furries''), catholics, progressives, leftists, white women, antifa, germans, journalists, islamists, southerners, media, religious people, assault victims, mass shooting victims, terrorism victims, ugly folks, atheist, buddhist, mormon, specific country, teenagers, young\_adults, terrorism, humanity, left\_wing\_people, terrorists, mormons, atheists, young adults, nonreligious people\\
\hline
\end{tabular}
\caption{Labels of relative social power assigned to lowercased identity terms from hate speech datasets.
Any identities not in these lists were marked \textit{other} by default.}
\label{tab:marginalized_dominant_identities}
\end{table*}

We applied the following transformations to datasets for binary hate speech labels:
\begin{itemize}
\item Civil Comments~\citep{Borkan2019}: toxicity value >= 0.5 was labeled hate
\item Social Bias Inference Corpus~\citep{Sap2020}: offensive value > 0.5 was labeled hate, following the original paper's binarization
\item \citet{Kennedy2020}: hate speech value > 1 was labeled hate
\item HateXplain ~\citep{mathew2021hatexplain}: labeled hate if any annotator labeled the instance as hate
\item Contextual Abuse Dataset~\citep{vidgen-etal-2021-introducing}: labeled hate if any of the following labels was present: \texttt{AffiliationDirectedAbuse}, \texttt{Slur}, \texttt{IdentityDirectedAbuse}
\item \citet{elsherief2021latent}: we paired implicit hate (which was annotated with identity targets) with non-hate from stage 1 annotations
\item \citet{salminen2018anatomy}: labeled hate if the class was labeled \texttt{hateful}
\end{itemize}

\end{document}